\def\year{2022}\relax
\documentclass[letterpaper]{article} 
\usepackage{aaai22}  
\usepackage{times}  
\usepackage{helvet}  
\usepackage{courier}  
\usepackage[hyphens]{url}  
\usepackage{graphicx} 
\usepackage{natbib}  
\usepackage{caption} 
\DeclareCaptionStyle{ruled}{labelfont=normalfont,labelsep=colon,strut=off} 
\usepackage{algorithm}
\usepackage{algorithmic}
\usepackage{bbding}
\usepackage{amssymb}
\usepackage{amsmath}
\usepackage{pifont}
\newcommand{\cmark}{\ding{51}}%
\newcommand{\xmark}{\ding{55}}%
\usepackage{multirow}
\usepackage{newfloat}
\usepackage{listings}
\usepackage[colorlinks, linkcolor=blue]{hyperref}
\floatstyle{ruled}
\newfloat{listing}{tb}{lst}{}
\floatname{listing}{Listing}
\title{QueryProp: Object Query Propagation for High-Performance Video Object Detection}
\author{
Fei He\textsuperscript{\rm 1,2}, 
Naiyu Gao\textsuperscript{\rm 1,2}, 
Jian Jia\textsuperscript{\rm 1,2}, 
Xin Zhao\textsuperscript{\rm 1,2}\thanks{Corresponding author}, 
Kaiqi Huang\textsuperscript{\rm 1,2,3}
}
\affiliations{
\textsuperscript{\rm 1}CRISE, Institute of Automation, Chinese Academy of Sciences\\
\textsuperscript{\rm 2}School of Artificial Intelligence, University of Chinese Academy of Sciences \\
\textsuperscript{\rm 3}CAS Center for Excellence in Brain Science and Intelligence Technology \\
\{hefei2018, gaonaiyu2017, jiajian2018\}@ia.ac.cn, \{xzhao, kaiqi.huang\}@nlpr.ia.ac.cn 
}

\begin{document}

\maketitle

\begin{abstract}
Video object detection has been an important yet challenging topic in computer vision.
Traditional methods mainly focus on designing the image-level or box-level feature propagation strategies to exploit temporal information. 
This paper argues that with a more effective and efficient feature propagation framework, video object detectors can gain improvement in terms of both accuracy and speed.
For this purpose, this paper studies object-level feature propagation, and proposes an object query propagation (QueryProp) framework for high-performance video object detection.
The proposed QueryProp contains two propagation strategies:
1) query propagation is performed from sparse key frames to dense non-key frames to reduce the redundant computation on non-key frames;
2) query propagation is performed from previous key frames to the current key frame to improve feature representation by temporal context modeling.
To further facilitate query propagation, an adaptive propagation gate is designed to achieve flexible key frame selection. 
We conduct extensive experiments on the ImageNet VID dataset. 
QueryProp achieves comparable accuracy with state-of-the-art methods 
and strikes a decent accuracy/speed trade-off. 
Code is available at \href{https://github.com/hf1995/QueryProp}{https://github.com/hf1995/QueryProp}.
\end{abstract}

\section{Introduction}
\begin{figure}[t]
    \centering
    \includegraphics[width=0.95\columnwidth]{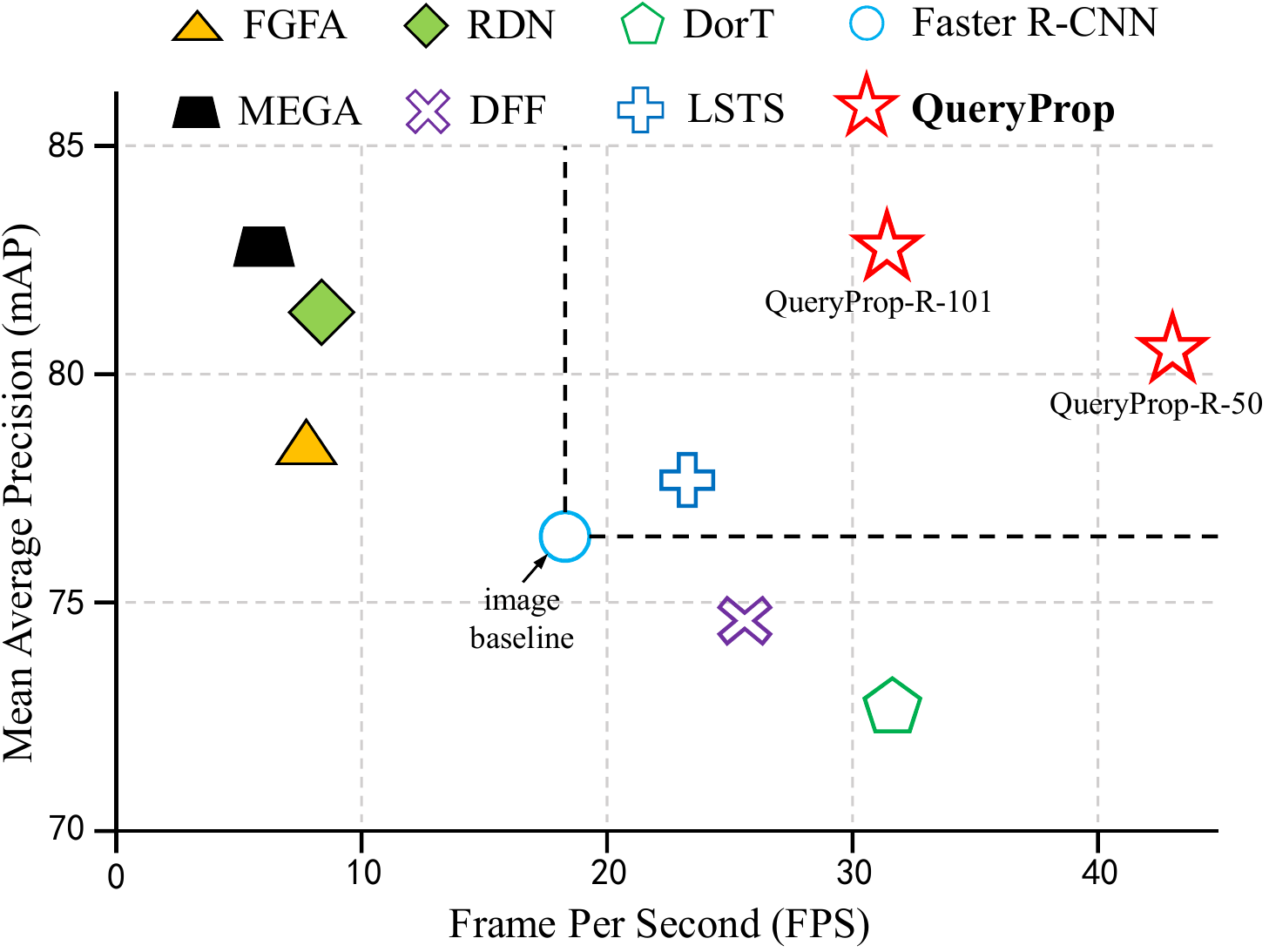} 
    \caption{mAP \textit{vs.} FPS on ImageNet VID dataset. The hollow/solid symbol represents online/offline algorithms. 
            Methods include FGFA~\cite{zhu17fgfa}, MEGA~\cite{chen2020memory}, RDN~\cite{deng19rdn}, 
            DFF~\cite{zhu17dff}, DorT~\cite{luo2019dort}, LSTS~\cite{jiang2020learning}, Faster R-CNN~\cite{ren15faster} and our proposed QueryProp. 
            QueryProp outperforms all previous online video detectors and strikes a decent accuracy/speed trade-off. 
            All methods are tested on a TITAN RTX GPU.}
    \label{map_vs_fps}
    \vspace{-1.4em}
\end{figure}

Object detection is one of the most fundamental and challenging problems in computer vision. 
Over the past decades, remarkable breakthroughs have been made in object detection, with the advances of well-designed modules, e.g., anchor generator and NMS~\cite{ren15faster, liu16ssd, tian2019fcos}. 
Recently, DETR~\cite{carion2020detr} and follow-up works~\cite{zhu2020defdetr,sun2021sparse} reformulate object detection as a query-based set prediction problem to simplify the detection pipeline. 
These query-based detectors no longer require heuristic components and have achieved comparable performance with state-of-the-art methods. 

Although promising results can be achieved on static images, it might not be suitable to directly extend image detectors to videos. 
The reasons mainly lie in two aspects. 
First, object appearances are usually deteriorated by motion or severe occlusion in videos, 
which bring extreme challenges to image detectors.
Second, existing image detectors generally depend on cumbersome stacked modules to generate accurate predictions. 
It might be inefficient to directly perform frame-by-frame prediction on videos using image detectors, 
and the computation burden also restricts the applications of detectors in real-world systems~\cite{huang2017speed}.

To address the above challenges, the core idea is to exploit temporal context in videos to boost detection performance. 
Figure~\ref{map_vs_fps} visualizes recent progress of video detectors on ImageNet VID dataset~\cite{deng2009imagenet}.
As is illustrated, most state-of-the-art detectors gain performance improvement in terms of either accuracy or speed. 
That is, detection accuracy is usually promoted at the expense of speed, and vice versa.
Specifically, to obtain more accurate predictions of deteriorated frames, 
some methods~\cite{zhu17fgfa,deng19rdn,chen2020memory} adopt dense feature aggregation strategy for feature representation. 
In these methods, each frame is enhanced in an undifferentiated way by aggregating features from its multiple adjacent frames, 
and additionally brings excessive computation cost compared with image detectors (e.g., Faster R-CNN~\cite{ren15faster}). 
To speed up video detectors, some methods~\cite{zhu17dff,zhu18towards,jiang2020learning} are designed to exploit temporal continuity among video frames for acceleration. 
In these methods, features of non-key frames are computed by only propagating features from sparse key frames, 
therefore avoid repetitive dense feature extraction operations on each frame. 
However, it is noticeable that additional feature alignment modules are necessary for image-level feature propagation due to the frame variations.
Besides, the prediction error of spatial offsets between features of adjacent frames may cause representation degradation, and then leads to an accuracy decrease.
By far, it remains a challenge to obtain an ideal accuracy/speed trade-off for video detectors.

Inspired by recent success of query-based image detectors~\cite{carion2020detr,zhu2020defdetr,sun2021sparse}, 
this paper proposes an object query propagation (QueryProp) framework to achieve efficient and effective video object detection. 
In image detectors, object queries encode instance information of each frame. 
Each object query is randomly initialized and a multi-stage structure is then constructed to iteratively update object features. 
In videos, the same objects usually appear with high probability at nearby positions between consecutive frames, so that the redundant iteration process can be further simplified based on video continuity.
To achieve more efficient detection, our proposed QueryProp propagates object queries from key frames to nearby non-key frames for query initialization instead of random initialization. 
Therefore, accurate detection results can be obtained with much fewer refine modules on non-key frames. 
In the meanwhile, QueryProp propagates the object queries in the previous key frames to the current key frame, which explores temporal relations between object queries to further enhance feature representation. 
Different from traditional methods that utilized intuitive key frame selection strategy, 
an adaptive propagation gate (APG) is proposed to flexibly select key frames.
The APG module estimates the reliability of the query propagation results, and automatically determines whether to select the current frame as a key frame.
In our proposed method, detection loss is adopted to measure query propagation quality and generate pseudo-labels to train the APG module in a self-supervised manner.
Compared with traditional methods, the proposed QueryProp can perform effective feature propagation using a more lightweight module.

The main contributions of this paper lie in three aspects: 
\begin{itemize}
    \item This paper is the first to propose a query-based propagation (QueryProp) method for video object detection.
    \item Two effective feature propagation strategies are proposed to simultaneously simplify redundant refine the structure and enhance feature representations, together with the proposed APG module for adaptive key frame selection.
    \item Experiments on the ImageNet VID demonstrate that QueryProp achieves comparable accuracy with the state-of-the-art video object detectors with a much faster speed. 
\end{itemize}

\section{Related Work}
\subsection{Image Object Detection}
Object detection has achieved remarkable results on static images. 
Mainstream detectors are anchor-based methods, which include one-stage and two-stage detectors.
One-stage detectors directly predict the offsets of the preset anchors to get the final results. 
Related works include YOLO~\cite{redmon16yolo}, SSD~\cite{liu16ssd}, RetinaNet~\cite{retinanet}, etc. 
Two-stage detectors first regress the anchors to generate the candidate regions, and then classify and regress the RoI features for final prediction. 
Related works include R-CNN~\cite{rbg14rcnn}, Fast R-CNN~\cite{rbg15fast}, and Faster R-CNN~\cite{ren15faster}, etc. 
FCOS~\cite{tian2019fcos} and CenterNet~\cite{duan2019centernet} establish anchor-free detectors with competitive detection performance. 
Recently, DETR~\cite{carion2020detr} reformulates object detection as a query-based set prediction problem and receives lots of attention. 
Deformable DETR~\cite{zhu2020defdetr} introduces deformable attention module to DETR for efficiency and fast-convergence. 
Sparse R-CNN~\cite{sun2021sparse} builds a query-based detector on top of R-CNN architecture with learnable proposal boxes and proposal features. 

\begin{figure*}[t]
  \centering
  \includegraphics[width=0.95\textwidth]{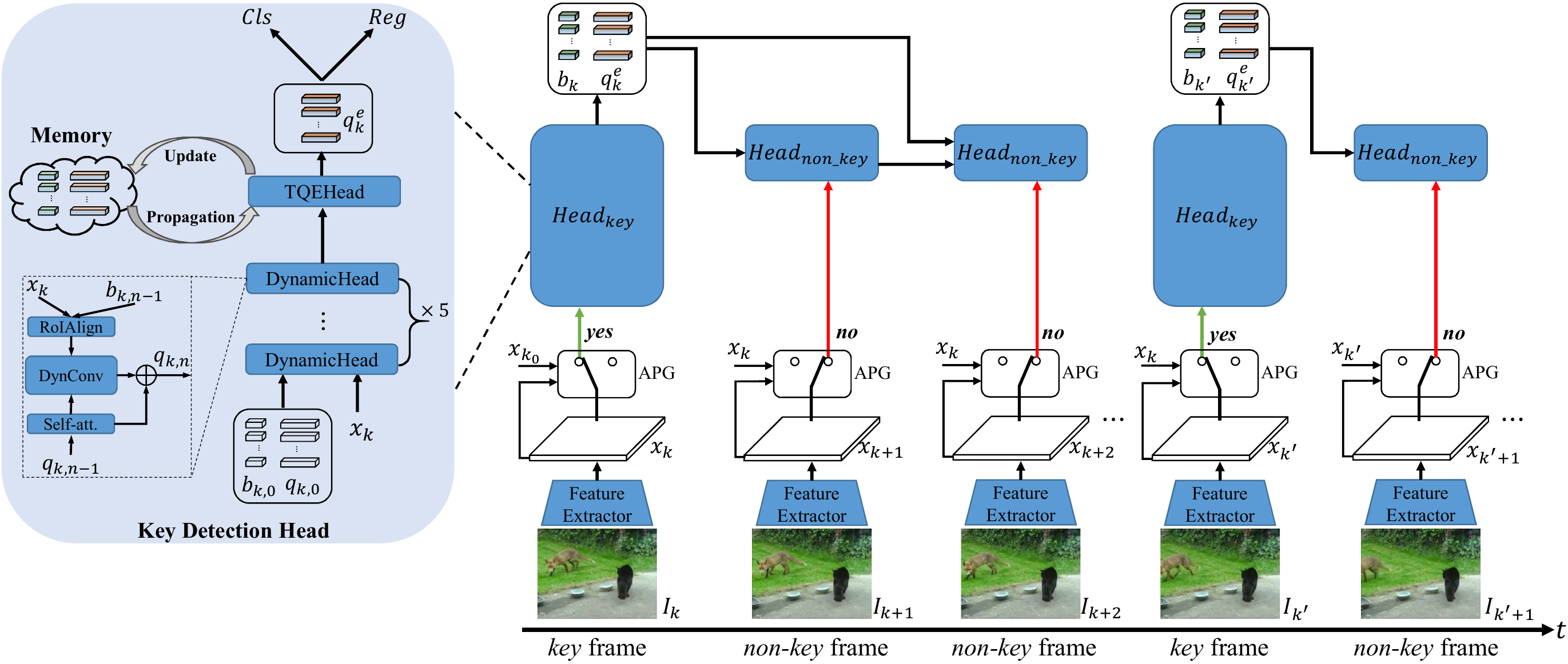} 
  \caption{Overview of QueryProp. 
            Given a video, at time step $k$, frame $I_{k}$ is fed to feature extractor to obtain features $\boldsymbol{x}_{k}$. 
            Then, the previous key frame features $\boldsymbol{x}_{k_0}$ and $\boldsymbol{x}_{k}$ are fed to the adaptive propagation gate (APG), 
            which evaluates whether to set $I_{k}$ as a new key frame. 
            The decision is $yes$, and $I_k$ is set as a key frame. 
            $\boldsymbol{x}_{k}$ is fed to the key detection head (five dynamic heads and one temporal query enhance head) to generate detection results. 
            At time step $k+1$, the previous key frame features $\boldsymbol{x}_{k}$ and $\boldsymbol{x}_{k+1}$ are fed to APG.            
            The decision is $no$, $\boldsymbol{x}_{k+1}$ is fed to the non-key detection head to quickly get detection results.   }
  \label{framework}
  \vspace{-1.0em}
\end{figure*}
 \subsection{Video Object Detection}
 Videos contain rich yet redundant temporal information, 
 which makes video object detection and image object detection quite different. 
 Video object detection leverages temporal information to boost detectors generally in two directions, efficiency, and accuracy. 

 To improve the efficiency of video object detection, the main idea is to propagate image-level features across frames to avoid dense feature extraction. 
 DFF~\cite{zhu17dff} runs the expensive feature extractor on sparse key frames. 
 The features of key frames are warped to non-key frames with optical flow~\cite{flownet17}, 
 thus reducing computation cost and accelerating the whole framework. 
 Impression network~\cite{hetang2017impression} introduces sparse recursive feature aggregation to DFF for improving performance. 
 MMNet~\cite{wang2019fast} utilizes motion vector embedded in the video compression format to replace optical flow. 
 LSTS~\cite{jiang2020learning} and PSLA~\cite{guo2019progressive} propose well-designed sampling templates to achieve feature warping. 
 THP~\cite{zhu18towards} designs spatially adaptive feature updating and key-frame selection mechanisms to improve accuracy as well as speed. 
 However, the per-pixel motion estimation process is error-prone and slow, which is the bottleneck for higher performance. 
 ST-Lattice~\cite{chen2018lattice} propagates bounding boxes on key frames to non-key frames according to motion and scales. 
 DorT~\cite{luo2019dort} uses a tracker to propagate bounding boxes across frames.

 To improve the detection accuracy in degenerated frames, some post-processing methods have emerged in the early stage, 
 which are usually achieved by bounding box association investigation, 
 e.g., Seq-NMS~\cite{han2016seqnms}, T-CNN~\cite{kang2017tcnn}, \cite{kang2016object}, D\&T~\cite{feichtenhofer2017detect}. 
 The recent major solution is to aggregate features from nearby frames to leverage spatial-temporal coherence for feature enhancement. 
 FGFA~\cite{zhu17fgfa} applies optical flow to align features and aggregates the aligned features for feature enhancement. 
 Based on FGFA, MANet~\cite{wang18fully} adds an instance-level feature alignment module besides the pixel-level feature alignment. 
 STSN~\cite{gedas18stsn}, TCENet~\cite{he20tcenet} and TASFA~\cite{he2022PR} apply deformable convolution~\cite{dai2017dcn} to perform frame-by-frame spatial alignment. 
 Instead of adopting a dense aggregation strategy, TCENet and TASFA propose stride predictors to adaptively select valuable frames to aggregate.  
 STMN~\cite{xiao2018stmm} devises a MatchTrans module to achieve feature alignment and aggregates features with well-designed recurrent units. 
 SELSA~\cite{wu19selsa} and LLRTR~\cite{shvets19llrtr} use similarity measures to aggregate proposal features to enhance object features.  
 RDN~\cite{deng19rdn}, OGEMN~\cite{deng2019object}, and MEGA~\cite{chen2020memory} 
 utilize relation network~\cite{hu2018relation} to model object relation in video to augment object features. 
 HVR-Net~\cite{han2020mining} uses inter-video and intra-video proposal relation to improve object feature quality. 
 However, the above methods require multi-frame aggregation and run at a slow speed, which is unable to meet the requirements in real-time systems.  

\section{Method}

In this section, we first briefly introduce the related query-based image detector. 
Then we describe the overall architecture of QueryProp, which consists of 
$1)$ object query propagation for computation acceleration, 
$2)$ object query propagation for query enhancement, 
and $3)$ adaptive propagation gate for flexible key frame selection. 

\subsection{Query-based Image Detector}
QueryProp is built on a multi-stage query-based image detector. 
Considering the performance and memory consumption of the detector, we build our method based on Sparse R-CNN~\cite{sun2021sparse}, 
which has six dynamic heads by default. 
Sparse R-CNN has $N$ learnable boxes $\boldsymbol{b}_{k,0}$ and queries $\boldsymbol{q}_{k,0}$, which are iteratively updated to fuse object features. 
At stage $n$, the dynamic head is shown in Figure~\ref{framework} and the pipeline can be formulated as follows: 
\begin{equation}
    \begin{aligned}
    \boldsymbol{x}_{k,n}^{\mathtt{roi}}    &= \mathrm{RoIAlign}(\boldsymbol{x}_k, \boldsymbol{b}_{k,n-1}), \\
    \boldsymbol{q}_{k,n-1}^*  &= \mathrm{SelfAtt}(\boldsymbol{q}_{k,n-1}),    \\
    \boldsymbol{q}_{k,n}      &= \mathrm{DynConv}(\boldsymbol{x}_{k,n}^{\mathtt{roi}}, \boldsymbol{q}_{k,n-1}^*) + \boldsymbol{q}_{k,n-1}^*,   \\
    \boldsymbol{b}_{k,n}      &= \mathrm{Reg}(\boldsymbol{q}_{k,n}),
    \end{aligned}
\end{equation}
where $\boldsymbol{x}_k$ is feature map, $\boldsymbol{b}_{k,n-1}$ and $\boldsymbol{q}_{k,n-1}$ are boxes and queries from the previous stage. 
Dynamic head utilizes RoIAlign~\cite{he2017mask} to extract RoI features $\boldsymbol{x}_{k,n}^{\mathtt{roi}}$, 
and $\boldsymbol{q}_{k,n-1}$ is processed by a self-attention module to generate proposal features $\boldsymbol{q}_{k,n-1}^*$. 
Then, a well-designed dynamic convolution module takes $\boldsymbol{q}_{k,n-1}^*$ and $\boldsymbol{x}_{k,n}^{\mathtt{roi}}$ as inputs to generate $\boldsymbol{q}_{k,n}$. 
Finally, $\boldsymbol{q}_{k,n}$ is fed into the box regression branch for box prediction $\boldsymbol{b}_{k,n}$. 

\subsection{QueryProp Architecture}
We propose QueryProp, an object query propagation framework for high-performance video object detection. 
Our goal is to reduce the overall computational cost of video detection while maintaining competitive accuracy in degenerated frames. 
The main idea of QueryProp is to leverage the strong continuity among consecutive frames to carry out efficient cross-frame object query propagation. 
Figure~\ref{framework} illustrates the overall architecture of QueryProp. 
Given a video, at each time step $k$, frame $I_k$ is fed to feature extractor to generate features $\boldsymbol{x}_k$. 
By considering the consistency between the previous key frame features $\boldsymbol{x}_{k_0}$ and $\boldsymbol{x}_k$, 
APG evaluates whether to set $I_k$ as a new key frame. 
The decision is $yes$, and $I_k$ is set as a key frame. 
$\boldsymbol{x}_k$ is fed to the key detection head, 
which consists of five dynamic heads and one temporal query enhance head~(TQEHead). 
In the TQEHead, queries of current frame aggregate queries from previous key frames to improve the quality of query features. 
The enhanced queries are used to generate final detection results and propagated to subsequent non-key frames. 
At time step $k+1$, the previous key frame features $\boldsymbol{x}_k$ and $\boldsymbol{x}_{k+1}$ are fed to APG. 
The decision is $no$, and $\boldsymbol{x}_{k+1}$ is fed to the lightweight non-key detection head. 
Queries and boxes from the previous key frame are utilized to initialize the current detection head. 
And detection results can be quickly generated without an accuracy decrease. 
Hence, by flexibly selecting key frames and efficiently propagating object queries between continuous frames, 
QueryProp can significantly reduce the overall computation cost and maintain competitive accuracy.

\subsection{Propagation for Computation Acceleration}
\begin{figure}[t]
  \centering
  \includegraphics[width=0.78\columnwidth]{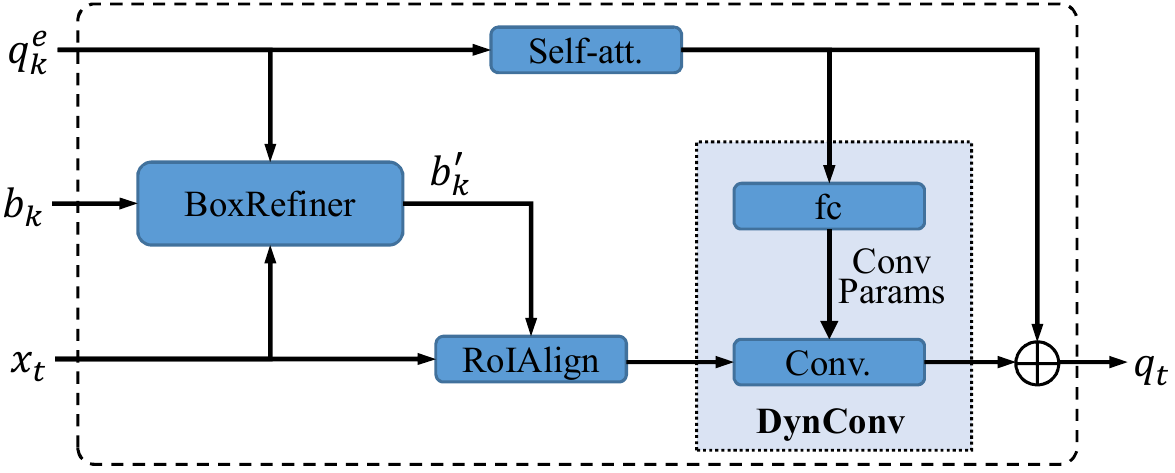} 
  \caption{Illustration of the non-key detection head. The boxes $\boldsymbol{b}_k$ of the previous key frame are first refined to generate more accurate object regions $\boldsymbol{b}'_k$. 
            Then the previous key frame queries $\boldsymbol{q}_k^e$ interact with RoI features of $\boldsymbol{b}'_k$ through a dynamic convolution to generate $\boldsymbol{q}_t$, which is used for final prediction.
           }
  \label{non_key_head}
  \vspace{-0.8em}
\end{figure}
To speed up video object detectors, 
the main idea of previous methods is to exploit temporal context among consecutive frames to reduce redundant computations. 
Specifically, the features on the dense non-key frames are obtained by propagating features from sparse key frames, 
thus avoiding feature extraction on most frames. 
Due to the feature inconsistency between the image-level features of adjacent video frames, 
these image-level feature propagation methods require additional models for per-pixel feature alignment. 
For example, some methods~\cite{zhu17dff,zhu18towards} utilize motion flow based feature warping, 
and others~\cite{guo2019progressive, jiang2020learning} utilize learnable feature sampling. 
However, the feature alignment models that resolve the inconsistency between the features of adjacent frames are often insufficient, 
which leads to an accuracy decrease. 
We design an efficient cross-frame object query propagation method to accelerate computation. 
Specifically, according to the strong continuity among consecutive frames, 
the detection head of the current frame can be initialized with the object queries from the previous frame. 
Benefiting from the reasonable initialization, 
the detection results can be quickly obtained with a more lightweight structure instead of the commonly employed 6-stage structure.

If current frame $I_t$ is a non-key frame, features $\boldsymbol{x}_t$, as well as the queries $\boldsymbol{q}_k^e$ and boxes $\boldsymbol{b}_k$ of the previous key frame, 
are input into the non-key detection head to quickly obtain the detection results.  
The detection pipeline is illustrated in Figure~\ref{non_key_head}, 
and can be formulated as follows: 
\begin{equation}
    \begin{aligned}
    \boldsymbol{b}_k'                &= \mathrm{BoxRefiner}(\boldsymbol{x}_t, \boldsymbol{b}_k, \boldsymbol{q}_k^e),  \\
    \boldsymbol{x}_{k\to t}^{\mathtt{roi}}    &= \mathrm{RoIAlign}(\boldsymbol{x}_t, \boldsymbol{b}_k'), \\
    \boldsymbol{q}_{k}^{e*}          &= \mathrm{SelfAtt}(\boldsymbol{q}_{k}^e),    \\
    \boldsymbol{q}_{t}               &= \mathrm{DynConv}(\boldsymbol{x}_{k\to t}^{\mathtt{roi}}, \boldsymbol{q}_{k}^{e*}) + \boldsymbol{q}_{k}^{e*},   \\
    \boldsymbol{c}_t, \boldsymbol{b}_t            &= \mathrm{Cls}\& \mathrm{Reg}(\boldsymbol{q}_t).
    \end{aligned}
\end{equation}
The boxes $\boldsymbol{b}_k$ are first refined with a box refiner to generate more accurate object regions, 
and the corrected boxes $\boldsymbol{b}'_k$ are used to generate RoI features $\boldsymbol{x}_{k\to t}^{\mathtt{roi}}$. 
Each query in $\boldsymbol{q}_{k}^e\in \mathbb{R}^{N\times C}$ interacts with the corresponding RoI feature in $\boldsymbol{x}_{k\to t}^{\mathtt{roi}}\in \mathbb{R}^{N\times S\times S \times C}$ to 
filter out ineffective bins in a RoI and get the final object feature $\boldsymbol{q}_{t}$. 

Feature interaction is implemented by the dynamic convolution~\cite{sun2021sparse}.
Dynamic convolution consists of $1\times 1$ convolutions, and $\boldsymbol{q}_{k}^{e*}$ generates the convolutional kernel parameters. 
Particularly, the feature interaction can be seen as a spatial attention mechanism to focus on `where' is an important part in a RoI of size $S\times S$. 
$\boldsymbol{q}_{k}^{e*}$ generates the attention weights to indicate the most relative bins in a RoI for final object location and classification. 
There are two ways to implement the box refiner. 
The first one refines the box through a dynamic convolution, 
and the second refines the box through the detection heads between $I_k$ and $I_t$. 
We discuss these two implementations in the ablation study. 
The whole pipeline does not require other models except the detection modules.

\subsection{Propagation for Query Enhancement}
\begin{figure}[t]
  \centering
  \includegraphics[width=0.95\columnwidth]{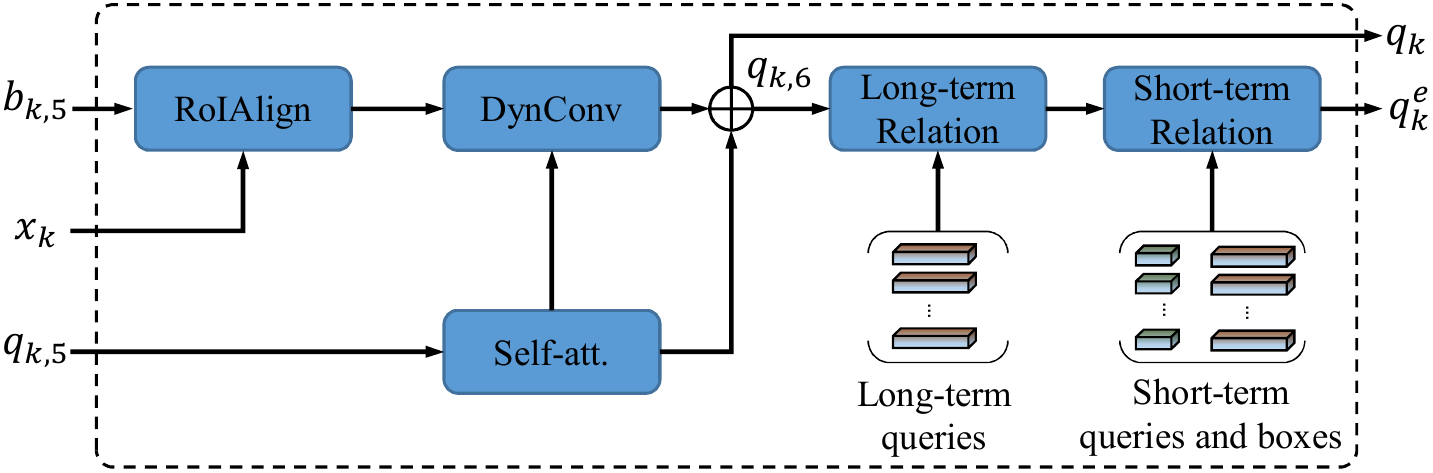} 
  \caption{Illustration of TQEHead in the key detection head. 
           Queries $\boldsymbol{q}_{k,5}$ from 5-$th$ stage are first updated to generate $\boldsymbol{q}_{k,6}$. 
           Then $\boldsymbol{q}_{k,6}$ separately aggregates long-term memory and short-term memory for feature enhancement. 
           $\boldsymbol{q}_{k,6}$ is sorted according to the confidence score to get $\boldsymbol{q}_k$, which is used to update the memory feature. }
  \label{key_head}
  \vspace{-0.8em}
\end{figure}
To improve detection accuracy of deteriorated frames, 
previous methods~\cite{zhu17fgfa,deng19rdn} usually adopt dense feature aggregation on all frames to improve feature quality, which is quite slow. 
Due to the similar appearance among adjacent frames, dense aggregation is sub-optimal. 
It is more efficient to aggregate features on sparse key frames for feature enhancement and propagate the enhanced features to dense non-key frames. 

Therefore, we design object query propagation between previous key frames and current key frame for query enhancement. 
As mentioned, if current frame is set as a new key frame, 
we denote its features by $\boldsymbol{x}_k$ and feed it to the key detection head. 
The randomly initialized queries and boxes, as well as $\boldsymbol{x}_k$, are input into five dynamic heads to iteratively update object features (as shown in Figure~\ref{framework}).
A temporal query enhance head~(TQEHead) then aggregates queries of previous key frames to model temporal relation for query feature enhancement. 
The details of the TQEHead are illustrated in Figure~\ref{key_head}. 
$\boldsymbol{q}_{k,5}$ from the 5-$th$ dynamic head is first updated to generate queries $\boldsymbol{q}_{k,6}$, 
which then aggregates with memory features for query feature enhancement. 
To enhance the quality of object features, 
we choose relation module~\cite{hu2018relation} to mine relations between object queries, 
which characterizes the interaction between their appearance feature and geometry. 
However, two objects that have long-term temporal distance may not have effective geometry relations. 
Therefore, we divide the queries in memory into two categories according to the distance of temporal dimension, 
long-term memory and short-term memory. 
The query enhancement process can be expressed as follows:
\begin{equation}
    \begin{aligned}
    \boldsymbol{q}_k^l       &= \mathrm{LongRelation}(\boldsymbol{q}_{k,6}, \{\mathbf{L}_q\}),  \\
    \mathbf{S}_{q}^{l}   &= \mathrm{LongRelation}(\mathbf{S}_q, \{\mathbf{L}_q\}), \\
    \boldsymbol{q}_{k}^{e}   &= \mathrm{ShortRelation}(\boldsymbol{q}_{k}^l, \{\mathbf{S}_q^l, \mathbf{S}_b\}),    \\
    \boldsymbol{c}_{k}, \boldsymbol{b}_k  &= \mathrm{Cls}\& \mathrm{Reg}(\boldsymbol{q}_k^e),   \\
    \boldsymbol{q}_k         &= \mathrm{Sort}(\boldsymbol{q}_{k, 6}, \boldsymbol{c}_k),  \\
             \mathrm{Upd}&\mathrm{ate}(\mathbf{L}_q, \mathbf{S}_q, \mathbf{S}_b, \boldsymbol{q}_k, \boldsymbol{b}_k).
    \end{aligned}
\end{equation}
All queries and boxes in $M$ adjacent key frames are grouped together to form the short-term memory $\{\mathbf{S}_q, \mathbf{S}_b\}$. 
Top $l$ queries with their scores in each frame of remaining key frames are grouped together, 
and we randomly select $T$ queries to form the long-term memory $\mathbf{L}_q$. 
We first use $\mathbf{L}_q$ to enhance both $\boldsymbol{q}_{k,6}$ and $\mathbf{S}_q$. 
Then, the enhanced short-term memory $\{\mathbf{S}_q^l, \mathbf{S}_b\}$ are used to enhance $\boldsymbol{q}_{k}^l$ to generate $\boldsymbol{q}_k^e$,  
which is used to generate final detection results and propagated to the next frame. 
Finally, $\boldsymbol{q}_{k,6}$ is sorted according to confidence score to get $\boldsymbol{q}_k$, which is used to update the memory. 

\subsection{Adaptive Propagation Gate}
An important step in our framework is to decide whether the current frame should be set as a new key frame. 
Some methods~\cite{zhu17dff,jiang2020learning} adopt fixed-rate key frame schedulers regardless of the irregular change of object over time. 
Others~\cite{zhu18towards} adopt the simple threshold way based on motion flow to select key frames, which cannot be optimized together with the detector. 

We propose an adaptive propagation gate~(APG) with a learnable gating unit to flexibly select key frames. 
When the query propagation is unreliable and hard to generate accurate results, 
APG sets current frame as a new key frame. 
The gating unit $g(\boldsymbol{x}_{k}, \boldsymbol{x}_t)\to \{0, 1\}$ receives current frame feature $\boldsymbol{x}_t$ and the previous key frame feature $\boldsymbol{x}_{k}$ as inputs. 
The gate first generates residual feature $\boldsymbol{r}_t$, which represents the difference between the current frame and the previous key frame features $\boldsymbol{x}_t - \boldsymbol{x}_{k}$. 
Then $\boldsymbol{r}_t$ is fed to a $3\times 3$ convolution, and $4\times 4$ adaptive average pooling is performed. 
The flattened resulting features are linearly projected and fed to a sigmoid function. 
The gate is lightweight and only requires cheap computation for per-frame evaluation.

During training, the parameters of the gate are learned in a self-supervised way by minimizing the binary cross-entropy between the gating outputs and pseudo labels $y_t^g$. 
For each frame $I_{z_0}$, we randomly select $m$ adjacent frames as key frames, denoted as $\{I_{z_1},\cdots,I_{z_m}\}$. 
The gate loss is: 
\begin{equation}
  \mathcal{L}_{\mathtt{gate}}=\frac{1}{m}\sum_{t=1}^{m}\mathrm{BCE}(g(\boldsymbol{x}_{z_t},\boldsymbol{x}_{z_0}),y_t^g).
\end{equation}
The queries from key frame $I_{z_t}$ are propagated to non-key frame $I_{z_0}$ to generate detection results, 
and the classification loss is denoted as $\mathcal{L}_{\mathtt{cls}}(\boldsymbol{x}_{z_t}, \boldsymbol{x}_{z_0}, \boldsymbol{c}_{z_0})$, where $\boldsymbol{c}_{z_0}$ is the groundtruth of frame $I_{z_0}$. 
We use the classification loss to generate the pseudo labels:
\begin{equation}
    y_t^g = \begin{cases}
    1 & \mathcal{L}_{\mathtt{cls}}(\boldsymbol{x}_{z_t}, \boldsymbol{x}_{z_0}, \boldsymbol{c}_{z_0}) > \epsilon_{z_0}, \\
    0 & else,
    \end{cases}
\end{equation}
where $\epsilon_{z_0}$ determines the maximum loss required to set a new key frame. 
A label $1$ indicates the current key frame can not provide enough information for the non-key frame, and a new key frame needs to be set. 
A label $0$ indicates the non-key detection head can generate a reliable prediction with the current key frame. 
We define $\epsilon_{z_0}$ as:
\begin{equation}
  \epsilon_{z_0}=\beta \min \{\mathcal{L}_{\mathtt{cls}}(\boldsymbol{x}_{z_t}, \boldsymbol{x}_{z_0}, \boldsymbol{c}_{z_0})\}_{t=1}^{m},
\end{equation}
where $\beta$ is a hyper-parameter that controls the trade-off between accuracy and computation costs. 
The smaller the $\beta$ is, the higher the frequency of key frame selection is.  

\begin{table*}[t]
  \setlength{\abovecaptionskip}{0pt}
  \setlength{\belowcaptionskip}{-1pt}
  \centering
  \small
  \begin{tabular}{l|c|c|c|c|c}
    \hline
    Methods                             & Online    & Backbone          & Base Detector     & mAP(\%)   & FPS    \\
    \hline
    \hline
    FGFA~\cite{zhu17fgfa}               &  \xmark   & ResNet-101        & R-FCN             & 76.3      & 7.1        \\
    MANet~\cite{wang18fully}            &  \xmark   & ResNet-101        & R-FCN             & 78.1      & 8.4        \\
    STSN~\cite{gedas18stsn}             &  \xmark   & ResNet-101+DCN    & R-FCN             & 78.9      & -     \\
    ST-Lattice~\cite{chen2018lattice}   & \xmark    & ResNet-101        & Faster R-CNN      & 79.6      & 20.0 (X)  \\
    SELSA~\cite{wu19selsa}              &  \xmark   & ResNet-101        & Faster R-CNN      & 80.2      & -        \\
    TCENet~\cite{he20tcenet}            &  \xmark   & ResNet-101        & R-FCN             & 80.3      & 11.0        \\
    LLRTR~\cite{shvets19llrtr}          &  \xmark   & ResNet-101        & Faster R-CNN      & 80.6      & -        \\
    RDN~\cite{deng19rdn}                &  \xmark   & ResNet-101        & Faster R-CNN      & 81.8      & 7.5       \\ 
    MEGA~\cite{chen2020memory}          &  \xmark   & ResNet-101        & Faster R-CNN      & 82.9      & 5.5        \\
    \hline
    Faster R-CNN~\cite{ren15faster}     & \cmark    & ResNet-101        & Faster R-CNN      & 77.0      & 19.0          \\
    Sparse R-CNN~\cite{sun2021sparse}   & \cmark    & ResNet-101        & Sparse R-CNN      & 77.3      & 18.0          \\
    DFF~\cite{zhu17dff}                 & \cmark    & ResNet-101        & R-FCN             & 73.1      & 23.0          \\
    THP~\cite{zhu18towards}             & \cmark    & ResNet-101+DCN    & R-FCN             & 78.6      & 13.0 (X)         \\
    PSLA~\cite{guo2019progressive}      & \cmark    & ResNet-101        & R-FCN             & 77.1      & 30.8 (V)      \\
    OGEMN~\cite{deng2019object}         & \cmark    & ResNet-101        & R-FCN             & 79.3      & 8.9 (X)     \\
    LSTS~\cite{jiang2020learning}       & \cmark    & ResNet-101        & R-FCN             & 77.2      & 23.0 (V)          \\
    \textbf{QueryProp}                  & \cmark    & ResNet-101        &  Sparse R-CNN     & 82.3      & 32.5 / 26.8 (X)        \\
    \textbf{QueryProp}                  &  \cmark   & ResNet-50         &  Sparse R-CNN     & 80.3      & 45.6 / 36.8 (X) \\
    \hline
  \end{tabular}\smallskip
  \caption{Comparison to the state-of-the-art methods on the ImageNet VID. 
  Without special marking, the runtime is tested on a TITAN RTX GPU. 
  X means TITAN X, and V means TITAN V. 
  DCN represents deformable convolution~\cite{dai2017dcn}.}
  \label{Compare}
\end{table*}

\section{Experiments}
\subsection{Dataset and Evaluation}
 We evaluate our model on the ImageNet VID~\cite{deng2009imagenet}, 
 which consists of 3862 training videos and 555 validation videos from 30 object categories. 
 All videos are fully annotated with the object bounding box, object category, and tracking IDs. 
 We report mean Average Precision (mAP) on the validation set as the evaluation metric. 
 
 Following the setting in~\cite{zhu17fgfa}, both ImageNet VID and ImageNet DET~\cite{deng2009imagenet} are utilized to train our model. 
 Since the 30 object categories in ImageNet VID are a subset of 200 categories in ImageNet DET, 
 the images from overlapped 30 categories in ImageNet DET are adopted for training. 

 \subsection{Implementation Details}

 \subsubsection{Training setup.}
 The proposed framework is implemented with PyTorch-1.7. 
 QueryProp utilizes AdamW~\cite{loshchilov2018decoupled} optimizer with weight decay 0.0001. 
 The whole framework is trained with 8 GPUs and each GPU holds one mini-batch. 
 The framework is trained in two stages. 
 We first train the backbone and the detection heads on both ImageNet DET and VID. 
 Given a key frame $I_k$ and a non-key frame $I_{i}$, 
 we randomly sample two frames from $\{I_t\}_{t=k-3\tau}^{k-\tau}$ for short-term memory generation, 
 and two frames from $\{I_t\}_{t=0}^{k}$ for long-term memory generation. 
 If they are sampled from DET, all frames within the same mini-batch are the same since DET only has images. 
 We use the parameters pre-trained on COCO~\cite{lin2014microsoft} for model initialization. 
 The training iteration is set to 90k and the initial learning rate is set to $2.5\times 10^{-5}$, 
 divided by 10 at iteration 65k and 80k, respectively.
 After finishing the first training stage, 
 we start training the APG on ImageNet VID. 
 For each non-key frame, we randomly select $m$ (10 by default) adjacent frames as key frames to form a training batch. 
 The gate is optimized in a self-supervised manner. 
 The initial learning rate is set to $10^{-4}$ and the total training iteration is 16k, 
 and the learning rate is dropped after iteration 8k and 12k. 
 The number of queries and boxes in the detection heads is 100 by default. 

 \subsubsection{Inference.}
 Given a video frame $I_t$, the short-memory saves the queries and boxes from the previous $M$~(10 by default) key frames. 
 The long-memory saves top $l$~(50 by default) queries of each remaining key frame and $T$~(500 by default) queries are randomly selected for feature enhancement. 
 The memories are updated after the detection of each key frame is completed. 

\subsection{Main Results}
Table~\ref{Compare} shows the comparison between QueryProp and other state-of-the-art methods. 
Online setting means that the current frame can only access the information from the previous frames during inference. 
QueryProp adopts an online setting, which is more suitable for practical application. 
According to the online or offline setting is used in the algorithm, the existing methods can be divided into two categories. 
Methods with offline settings usually have more excellent accuracy but slow speed, 
which is hard to meet the requirements in real-time systems. 
Current methods using online settings usually have poor accuracy. 
Although they are faster in processing time than offline algorithms, most methods still cannot achieve real-time detection. 
Most of them utilize image-level propagation for computation acceleration, 
e.g., optical flow based feature warping, learnable feature sampling. 
Such per-pixel processes are quite slow and error-prone, which is the bottleneck for higher performance. 
QueryProp propagates sparse object queries across video frames to achieve online video object detection, 
and no additional modules or post-processing are required. 
As shown in Table~\ref{Compare}, QueryProp achieves the best performance on both accuracy and speed among all online methods. 
And the accuracy of QueryProp is comparable with most offline methods. 
When equipped with a lightweight backbone, 
the processing speed of QueryProp can achieve 45.6 FPS while maintaining an accuracy of over 80 mAP. 
 
\subsection{Ablation Study}
\begin{table}[t]
  \setlength{\abovecaptionskip}{0pt}
  \setlength{\belowcaptionskip}{-1pt}
  \centering
  \small
  \begin{tabular}{c|c|c|c|c|c}
      \hline
  Methods           & (a)     &  (b)    &      (c)      &    (d)          &     (e)             \\
          \hline\hline
  key$\to$non-key   &         & \cmark  &               & \cmark          & \cmark              \\
  key$\to$key       &         &         &   \cmark      & \cmark          & \cmark              \\
  APG               &         &         &               &                 & \cmark              \\
  \hline
  mAP(\%)           & 77.3    &  77.2   & 82.3          &  81.8           & 82.3                \\
  \hline
  FPS               & 18      &  33     & 16.9          & 32              & 32.5                \\
  \hline
  \end{tabular}\smallskip
  \caption{Accuracy and runtime of different methods on ImageNet VID validation.
  }
  \label{Ablation}
\end{table}
\subsubsection{Effectiveness of QueryProp.}
To demonstrate the effect of the proposed components in QueryProp, 
we conduct extensive experiments to study how they contribute to the final performance, 
and the results are summarized in Table~\ref{Ablation}. 
Method (a) is the single-frame baseline Sparse R-CNN using ResNet-101. 
Method (b) adds query propagation from key frame to non-key frames with key frame interval $k=10$,
which boosts the speed from 18 FPS to 33 FPS and almost no decrease in accuracy. 
Under similar settings, the image-level propagation methods~\cite{zhu17dff,jiang2020learning} usually cause a significant accuracy drop. 
Method (c) adds query propagation between key frames to (a) with $k=1$, 
which becomes a dense feature aggregation method. 
Each video frame is processed as a key frame, and the accuracy is significantly increased while the speed is decreased. 
By adding both two propagation to (a) with $k=10$, the accuracy and speed can be significantly improved at the same time. 
After adding adaptive key frame selection to (d), the performance of the algorithm is further improved, 
which can not only maintain a fast speed, but also achieve comparable accuracy with dense feature aggregation method. 
The above results indicate the effectiveness of each component in QueryProp. 

\begin{figure}[t]
  \centering
  \includegraphics[width=0.95\columnwidth]{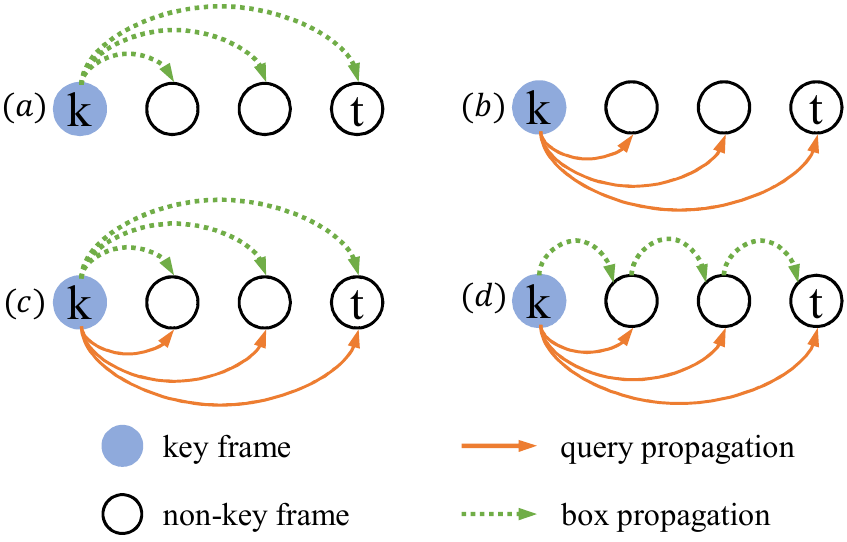} 
  \caption{Illustration of four different propagation architectures. }
  \label{prop_arch}
  
\end{figure}

\subsubsection{Propagation architecture design. }

\begin{table}[t]
  \setlength{\abovecaptionskip}{0pt}
  \setlength{\belowcaptionskip}{-1pt}
  \centering
  \small
  \begin{tabular}{c|c|c|c|c|c|c}
      \hline
  Methods   & baseline      & (a)       &  (b)      & (c1)  & (c2)  &  (d)            \\
          \hline\hline
      mAP(\%)  & 75.3       & 69.2      &  72.8     & 73.8  & 75.5  &  75.2        \\
  \hline
  FPS      & 22             & 36.4      &   36.4    & 44.5  & 36.4  & 44.5       \\
  \hline
  \end{tabular}\smallskip
  \caption{Accuracy and runtime of different propagation architecture in Figure~\ref{prop_arch}. Sparse R-CNN with ResNet-50 is adopted as baseline. }
  \label{arch_label}
  \vspace{-1.0em}
\end{table}

To verify the efficiency of object query propagation design, 
we compare four different propagation architectures shown in Figure~\ref{prop_arch}. 
Without special mentioning, we set key frame interval to 10 and adopt ResNet-50 as the backbone. 
Method (a) only propagates boxes of key frame $I_k$ to non-key frame $I_t$, 
while method (b) only propagates query features. 
Method (c) propagates both the boxes and queries of $I_k$ to $I_t$ at the same time. 
Particularly, (c1) removes the box refiner from the non-key detection head, 
and (c2) uses a dynamic convolution as the box refiner. 
Method (d) propagates the queries of $I_k$ to $I_t$, and the boxes are obtained from $I_{t-1}$. 
The non-key detection heads between $I_k$ and $I_t$ can be seen as the box refiner of $I_t$. 
The results are shown in Table~\ref{arch_label}. 
We can see query propagation is more important than box propagation, 
and better results can be achieved when the two propagation perform collaboratively. 
Method (c2) can achieve higher accuracy, but method (d) achieves a better accuracy/speed trade-off. 
We use the propagation structure of (d) by default.

\subsubsection{Adaptive vs. fixed key frame interval.}
\begin{table}[t]
  \setlength{\abovecaptionskip}{0pt}
  \setlength{\belowcaptionskip}{-1pt}
  \centering
  \small
  \begin{tabular}{c|c|c|c}
      \hline
  Methods                   & mAP(\%)   &  FPS      & Avg. interval    \\
          \hline\hline
  fixed (k=5)               & 82.2      &  28.6     & 5                \\
  fixed (k=10)              & 81.8      &  30.8     & 10              \\
  fixed (k=15)              & 81.5      &  33.1     & 15              \\
  \hline
  Adaptive ($\beta=1.25$)   & 82.5      &   29.6    & 6.6              \\                         
  Adaptive ($\beta=1.5$)    & 82.3      &   32.5    & 12.2              \\
  \hline
  \end{tabular}\smallskip
  \caption{Comparison of adaptive key frame selection and fixed selection. }
  \label{gate_compare}
\end{table}
We conduct an experiment to compare the adaptive key frame selection with the fixed key frame interval method. 
Table~\ref{gate_compare} shows the comparison results. 
When using a fixed key frame interval setting, accuracy is negatively correlated with interval while speed is positively correlated with interval. 
When using our adaptive selection setting, a better accuracy/speed trade-off is achieved. 
$\beta$ is a hyper-parameter in adaptive propagation gate, which controls the frequency of key frame selection. 
The larger $\beta$ is, the smaller the average interval is. 
We choose $1.5$ as the default value of $\beta$. 

\subsubsection{Ablation study on the memory. }
\begin{table}[t]
  \setlength{\abovecaptionskip}{0pt}
  \setlength{\belowcaptionskip}{-1pt}
  \centering
  \small
  \newcommand{\minitab}[2][l]{\begin{tabular}{#1}#2\end{tabular}}
  \begin{tabular}{c|cc|c}
    \hline 
    Methods    & $M$     & $T$     & mAP(\%)  \\
    \hline 
    \hline 
    \multirow{2}*{\minitab[c]{ QueryProp \\ (only short-term memory) }} & 10 & 0 & 79.1 \\
    & 20 & 0 & 79.4 \\
    \hline 
    \multirow{2}{*}{QueryProp } & 10 & 300 & 80.0 \\
    & 10 & 600 & 80.1 \\
    \hline
    \end{tabular}\smallskip
    \caption{Ablation study on the short-term and long-term memory. 
    }
  \label{memory_table}
\end{table}
Table~\ref{memory_table} shows the results of ablation study on short-term and long-term memory. 
The experiment is based on ResNet-50 and key frame interval of 10.  
To study the effect of long-term memory, we set the long-term memory size $L$ to 0 to remove its influence. 
As shown in the table, a significant drop in accuracy is obtained by removing the long-term memory. 
Gap still exists after increasing the short-memory size. 
The above results show the importance of long-term memory aggregation. 

\section{Conclusion}
This paper proposes a query-based high-performance video object detection framework, QueryProp, 
driven by efficient object query propagation between consecutive video frames. 
Specifically, QueryProp develops two propagation strategies to reduce redundant computation and improve the quality of object features.  
First, object queries from the sparse key frames are propagated to the dense non-key frames, 
which reduces the expensive computation on most frames. 
Second, object queries from the previous key frames are propagated to the current key frame, 
which exploits temporal information to improve the quality of query features. 
Besides, to enable efficient query propagation,  
QueryProp adopts an adaptive propagation gate to flexibly select key frames. 
Comprehensive experiments prove the effectiveness of our method. 
This novel solution enables such a new framework to achieve the best performance among all online video object detection approaches and strikes a decent accuracy/speed trade-off.

\section*{ Acknowledgments}
This work is supported in part by the National Natural Science Foundation of China (Grant No. 61721004), 
the Projects of Chinese Academy of Science (Grant No. QYZDB-SSW-JSC006), 
the Strategic Priority Research Program of Chinese Academy of Sciences (Grant No. XDA27000000), 
and the Youth Innovation Promotion Association CAS.

\bibliography{aaai22} 
\end{document}